\documentclass[lettersize,journal]{IEEEtran}
\usepackage{amsmath,amsfonts}
\usepackage{algorithmic}
\usepackage{array}
\usepackage[caption=false,font=normalsize,labelfont=sf,textfont=sf]{subfig}
\usepackage{textcomp}
\usepackage{stfloats}
\usepackage{url}
\usepackage{verbatim}
\usepackage{graphicx}
\def\BibTeX{{\rm B\kern-.05em{\sc i\kern-.025em b}\kern-.08em
    T\kern-.1667em\lower.7ex\hbox{E}\kern-.125emX}}
\usepackage{balance}
\begin{document}

\title{Enhancing Deep Learning based RMT Data Inversion using Gaussian Random Field}
\author{
    \IEEEauthorblockN{Koustav~Ghosal\IEEEauthorrefmark{1}, Arun~Singh*\IEEEauthorrefmark{1}, Samir~Malakar\IEEEauthorrefmark{2}, Shalivahan~Srivastava\IEEEauthorrefmark{3},
    Deepak~Gupta\IEEEauthorrefmark{4}\\}
    \IEEEauthorblockA{\IEEEauthorrefmark{1} Indian Institute of Technology (Indian School of Mines), Dhanbad, India\\}
    \IEEEauthorblockA{\IEEEauthorrefmark{2} UiT The Arctic University of Norway, Troms\o, Norway\\}
    \IEEEauthorblockA{\IEEEauthorrefmark{3} Indian Institute of Petroleum \& Energy, Visakhapatnam, India\\}
    \IEEEauthorblockA{\IEEEauthorrefmark{4} Transmute AI Lab (Texmin Hub), IIT(ISM), Dhanbad, India.}
}



\maketitle

\begin{abstract}
Deep learning (DL) methods have emerged as a powerful tool for the inversion of geophysical data. When applied to field data, these models often struggle without additional fine-tuning of the network. This is because they are built on the assumption that the statistical patterns in the training and test datasets are the same. To address this, we propose a DL-based inversion scheme for Radio Magnetotelluric data where the subsurface resistivity models are generated using Gaussian Random Fields (GRF). The network's generalization ability was tested with an out-of-distribution (OOD) dataset comprising a homogeneous background and various rectangular-shaped anomalous bodies. After end-to-end training with the GRF dataset, the pre-trained network successfully identified anomalies in the OOD dataset. Synthetic experiments confirmed that the GRF dataset enhances generalization compared to a homogeneous background OOD dataset. The network accurately recovered structures in a checkerboard resistivity model, and demonstrated robustness to noise, outperforming traditional gradient-based methods. Finally, the developed scheme is tested using exemplary field data from a waste site near Roorkee, India. The proposed scheme enhances generalization in a data-driven supervised learning framework, suggesting a promising direction for OOD generalization in DL methods.
\end{abstract}


\section{Introduction}
\IEEEPARstart{R}{adio} Magnetotelluric (RMT) method is one of the popular methods for near surface investigation. The application ranges from groundwater studies \cite{yogeshwar2012groundwater}, landslides \cite{shan2016integration}, study of waste sites \cite{wang2019boat}, etc. In this method, time-varying orthogonal components of electric and magnetic fields are measured on the Earth's surface. These fields are produced from the interaction of electromagnetic (EM) waves emitted from remote radio transmitters operating in the frequency range of 10-250 kHz and the Earth. From the recorded data, impedance (ratio of electric to magnetic field) is estimated. From the impedance data an estimate of the subsurface variation in electrical resistivity is obtained via inversion.

The gradient-based inversion methods such as occam inversion \cite{constable1987occam} etc, are the preferred optimization methods. These methods involve iteratively minimizing a penalty function defined as the sum of data misfit and model regularization. The solution obtained from these method depends on the choice of the initial model and model regularization term. Thus they fail to capture the full non-linearity between model and the observed data. Alternatively, global optimization such as simulated annealing \cite{shi1998one, sharma2012vfsares}, genetic algorithm \cite{schultz1993two, wang2018magnetotelluric}, ant colony algorithm \cite{liu2015two}, particle swarm optimisation algorithm \cite{shaw2007particle, shi2009damped} can be employed. These methods start with a random model and iteratively update it to minimize a predefined penalty function, typically representing data misfit. 
By exploring a broader search space and avoiding local minima, global optimization methods offer an alternative approach for inversion tasks. However, they are computationally very demanding, especially for a higher dimension problem.

Deep learning (DL) algorithms offer a promising solution for capturing the full non-linearity between data and model thus, addressing challenges encountered by traditional inversion algorithms. In recent years, DL techniques have seen extensive application in geophysics. \cite{liu2021deep} employed an 18-layer full Convolutional Neural Network (CNN) to perform 1D MT data inversion. Similarly, \cite{liao2022inversion} used a CNN combined with Long Short-Term Memory (LSTM) networks, while \cite{liu2023retrieval} utilized a 1D U-Net. \cite{ling2023one} adopted ResNet-18 and introduced parallel computation to expedite dataset construction. \cite{rahmani2024deep} conducted a comprehensive comparison of various network architectures for 1D MT data inversion. \cite{wang2022stochastic} employed reinforcement learning with a deep Q-network and transformed the inversion into a Markov decision process for uncertainty quantification using the Bayesian framework. \cite{rodriguez2023multimodal} utilized a multimodal variational autoencoder to generate multiple solutions and provided uncertainty quantification through truncated Gaussian densities, associating each solution with its respective uncertainty level. All the above studies are based on a supervised learning approach and need further fine-tuning for the inversion of field data.

For 2D MT data inversion, a deep belief network was utilized for supervised learning by \cite{wang2019nonlinear}, aiding in global search, however, the network was limited to a small-scale dataset.  \cite{guo2019regularized} proposed a method wherein the gradient descent direction was learned in the training phase, and the learned direction was utilized in the inference phase with regularized gradient descent. \cite{liu2021two} employed a deep belief network along with k-means++ for prior information clustering. Apart from these different deep convolutional networks have also been used for MT inversion. For example, Recurrent Neural Networks (RNNs) were utilized by \cite{jin2020rnn} to accelerate gradient calculation steps. A Fully Convolutional Network (FCN) was employed by \cite{liao20222d}, using datasets with simple geometric structures. As the number of frequency and grid points are different from the synthetic dataset, fine-tuning was required on additional test data. A ResNet with a dice loss function was utilized by \cite{xie20232d}, where the models had uniform backgrounds and simple geometric-shaped anomalous resistivity bodies. However, the network struggled with field data due to limited class diversity in the training dataset. A coupled U-Net based prior model with a regularized Gauss-Newton (GN) algorithm was proposed by \cite{wang2023flexible}, achieving good generalization. However, further training is required for field data.

The performance of data-driven DL methods depends heavily on the training data, as these methods assume that the training and test data share identical distributions. Previous studies have utilized simple resistivity models, such as rectangular-shaped structures and dipping structures embedded in a homogeneous background \cite{guo2019regularized, liu2021two, liao20222d, xie20232d}. To create more complex models, \cite{wang2023flexible} introduced irregular layered discontinuities with various polygon-shaped structures featuring anomalous resistivity values. However, these model representations still have a limited capacity to represent the complex resistivity distributions found in nature accurately. Such simplistic representation restricts the dataset's diversity, leading to poor generalization for out-of-distribution (OOD) samples. The OOD samples are statistically different from the training dataset.


Our study aims to enhance the generalizability of data-driven neural networks by focusing on two key aspects: (1) generating diverse datasets for training, (2) addressing missing data commonly observed in field datasets.
To achieve this, we utilize Gaussian Random Fields (GRFs) to create diverse resistivity distributions. GRFs are well-suited for simulating complex distributions and have been widely used across various disciplines, such as the study of ocean turbulent flow \cite{minakov2017acoustic}, flow in heterogeneous porous media \cite{wu20063d}, propagation of the seismic wave in heterogeneous media \cite{sato2012seismic}. In addition to generating diverse datasets, the issue of missing data is also addressed using a compressed sensing scheme \cite{candes2008introduction}. This method enables us to infer missing data points, thus enhancing the completeness of the dataset and improving the network's ability to learn from incomplete or sparse field data.

Compressed sensing(CS) is a powerful technique to reconstruct the original image or signal from a highly spare( or incomplete) image or signal sample. The basic requirement for CS is the sparsity of the signal and the incoherent nature of data acquisition. It has a wide range of applications from medical imaging\cite{graff2015compressive}, computer vision\cite{patel2013sparse}, remote sensing\cite{fan2016compressed}, etc. In addition to this, CS is also used for various geophysical data, such as time-limited time-lapse seismic data
reconstruction \cite{zhang2021time}, seismic data reconstruction\cite{gholami2014non}, seismic wavefield reconstruction\cite{muir2021seismic}. Application of CS other than seismic are spare aeromagnetic data was reconstruction \cite{o2024compressed} and CS-based 3D resistivity inversion \cite{ranjan2018compressed} which out-performs the existing algorithm in terms of accuracy and fast convergences with sparse data. This method enables us to infer missing data points, thus enhancing the completeness of the dataset and improving the network's ability to learn from incomplete or sparse field data.

The rest of the paper is organized as follows. Section \ref{method} gives a brief description of RMT forward modelling,  the inversion using DL, and the network architecture for the proposed inversion scheme. Section \ref{result} illustrates the training dataset, evaluating parameters, model training, and results of the different experiments with synthetic data. We test our method over the field data which is shown in section \ref{field_section}. We conclude our study in Section \ref{conclu} with scope for future developments.


\section{Method}
\label{method}
\subsection{RMT forward operator }
The computation of forward response for 2D RMT problem involves solving the transverse electric (TE-mode) and transverse magnetic (TM-mode) polarization. The governing partial differential equation for the two modes (assuming $y$ as the strike direction and time dependency of $e^{-\iota\omega t}$) is written as, 

\begin{equation}
    \label{TE_pde}
    \frac{\partial^2 E_x}{\partial y^2} + \frac{\partial^2 E_x}{\partial z^2} + i\omega\mu\sigma E_x = 0
\end{equation}
\begin{equation}
    \label{TM_pde}
    \frac{\partial }{\partial y}\Big(\rho\frac{\partial H_x}{\partial y} \Big) +
    \frac{\partial }{\partial z}\Big(\rho\frac{\partial H_x}{\partial z} \Big)
    - i \omega\mu H_x = 0
\end{equation}
In (\ref{TE_pde})-(\ref{TM_pde}), $E_x$ and $H_x$ are the $x$ component of electric and magnetic fields respectively, $\sigma$ is the electrical conductivity of the subsurface, $\rho$ is the electrical resistivity, $\epsilon$ is the electric permittivity, $\omega$ is the angular frequency of the EM wave and $\mu$ is the magnetic permeability. Once the electric (or magnetic) fields are computed, the magnetic field, $H_y$ (electric field, $E_y$) are evaluated. Using the electric and magnetic fields, the impedance are computed as $Z_{TE} = E_x/H_y$ and $Z_{TM} = E_y/H_x$. From these impedance values, apparent resistivity and phase values are obtained as, 
\begin{equation}
    \rho^{i}_a = \frac{|Z_{i}|^2}{\omega\mu}; ~~~~~~~~ \Phi^{i} = \arctan \Big( \frac{Im\big(Z_{i}\big)}{Re\big(Z_{i}\big)}\Big)
    \label{starteq}
\end{equation}
where $i$ represents the TE and TM mode. In this work, the forward responses are simulated using a Finite difference scheme implemented by \cite{singh2014interpretation}. 

\subsection{Deep Convolutional Network}
\label{workflow}
\begin{figure}[bt]
    \centering
    \includegraphics[width=0.8\columnwidth]{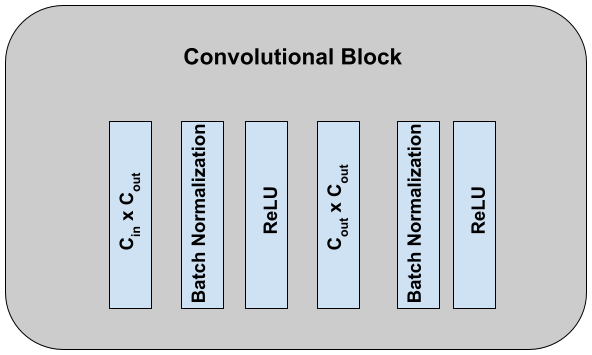}
    \caption{A schematic representation of a Convolutional block for each convolutional layer.}
    \label{fig:convolution_block}
\end{figure}

\begin{figure*}[bt]
    \centering
    \includegraphics[width=1.6\columnwidth]{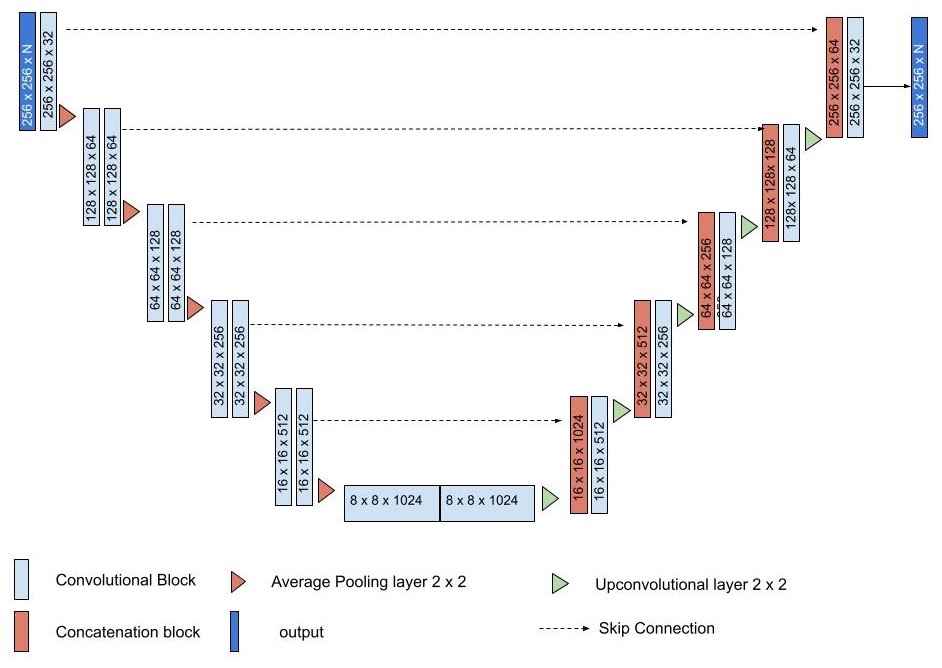}
    \caption{Image showing the proposed network for the RMT data inversion. The network consists of a sequence of convolutional layers, a batch normalization layer and a ReLu activation layer. The input is a 4 channel data (apparent resistivity and phase for TE and TM mode) and output is the resistivity model.}
    \label{fig:U-Net}
\end{figure*}

\noindent For the inversion of 2D RMT data, we have used a CNN-based encoder-decoder network known as U-Net \cite{ronneberger2015u}. U-Net is a U-shaped CNN network primarily used for image segmentation \cite{ronneberger2015u}. The proposed network design consists of a convolutional module (Fig. \ref{fig:convolution_block}) and consists of a sequence of convolutional layers, a batch normalization layer and a Rectified Linear Unit (ReLu) activation layer. After two consecutive convolutional modules and a 2D average pool layer, the input dimension changes from $(N, C, H, W)$ to $(N, 2C, \frac{H}{2}, \frac{W}{2})$ where $N$ is the batch size, $C$ is the number of channels, $H$ and $W$ are the height and width of the input (feature map) respectively. The bottleneck layer is defined with 512 channels and $16 \times 16$ dimensions of the feature map. The transpose convolution (or upconvolution) layer (kernel size=3, stride=2, padding=0, output padding=1) changes the dimension of the feature map from $(N, C, H, W)$ to $(N, \frac{C}{2}, 2H, 2W)$. The resulting image contains features from both high-level and low-level information. The U-net (see Fig. \ref{fig:U-Net}) was modified by incorporating following features
\begin{enumerate}
    \item \textit{Pooling Layer}: As the resistivity values in the models can take arbitrarily high and low values we use average pooling instead of max pooling as the average pooling smooths out the image and works better for our data.    
    \item \textit{Convolutional layer}: Instead of the single convolutional layer we introduced a convolutional block which consists of a sequence of convolutional layer, batch normalization layer, and ReLu activation layer.    
    \item \textit{Loss Function}: 
    We have used Mean Square Error (MSE) as the loss function since we have used U-Net for a regression task.
    \begin{equation}
    \label{lossEq}
        \textsc{Loss} = \frac{1}{N} \Big[\sum_{i,j} (\rho_a^{i,j} - \rho^{i,j})^2\Big]
    \end{equation}
where, $\rho_a^{i,j}$ and $\rho^{i,j}$ is the predicted and true resistivity respectively at a location ${i,j}$ in the subsurface. 
\end{enumerate}
The input for the network has 4 channels ($\rho^{TE}_a$, $\rho^{TM}_a$, $\Phi^{TE}$, $\Phi^{TM}$) and the output is a one-channel resistivity model. 

\section{Data Augmentation}
\subsection{Strategy of generating resistivity model using Gaussian Random Field}
\label{dataset_description}
The GRF, ${\mathit{F(x)}}$ defined on a domain $D \subset \mathbb{R}^d$ is a collection of random variables, where any finite set has a multivariate normal distribution, defined by its mean function $\mu(x)$ and covariance function $\mathit{C(x,y)}$. The Karhunen-Lo{è}ve expansion provides a series representation for such a GRF with zero mean (or after subtracting the mean) as 
\begin{equation}
    F(x) = \sum_{i=1}^{N} \sqrt{\lambda_i}\phi_i(x)\xi_i
\end{equation}
where $\lambda_i$ are the eigenvalues of the covariance operator, $\phi_i(x)$ are the corresponding eigenfunctions, and $\xi_i$ $\sim \mathcal{N}(0,1)$ are independent standard normal random variables. For a given Gaussian correlation function with spatial points $(x_1, x_2, ..., x_n)$ the correlation between points $x_i$ and $x_j$ is defined as
\begin{align}
    \textsc{corr}(x_i, x_j) &= \textsc{exp}((-1/2) \sum_i (( x_i - x_j )^2 / c_0(i))) \\
\end{align}
where $c_0$ is the correlation length. The top $k$ eigenvalues and eigenvectors used for generating the random field, are obtained from eigenvalue decomposition.
\begin{equation}
    [U, S] = \text{eig}(C)
\end{equation}
where $U$ contains eigenvectors $\phi_i$ and $S$ contains eigenvalues $\lambda_i$.
The generated random field realizations with the standard normal random variable $W$ are given as,
\begin{align}
    W &\sim \mathcal{N}(0,I_k) \\
    F &= U[:, 1:k] \sqrt{S[1:k, 1:k]} W + \mu
\end{align}
This process ensures that the generated random fields $F$ adhere to the desired mean and covariance structure specified by the correlation function and number of spatial points (mesh). To create diverse resistivity models (or realizations), the parameters of the correlation function were randomly selected within specified lower and upper bounds. Both isotropic and anisotropic correlations were considered, represented by scalar and vector scaling parameters, respectively. The polynomial expansion of the correlation function was truncated randomly between 5 and 10. The logarithmic resistivity values varied within a range of 1 to 4.



\begin{figure}[!h]
\centering
\includegraphics[width=\columnwidth]{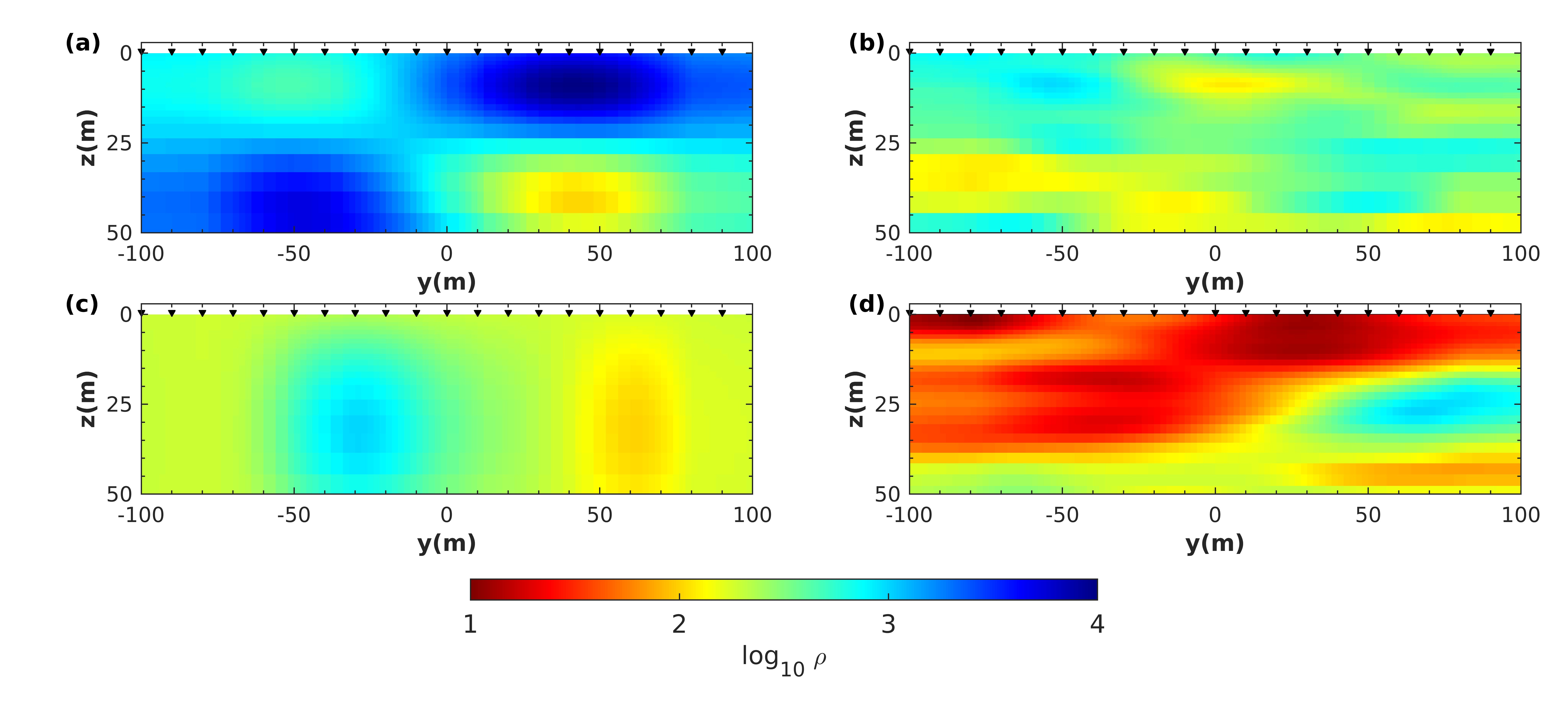}
\caption{The synthetic GRF dataset. (a) and (c) represent random localized anomalies, (b) and (d) represent randomly distributed horizontal or nearly horizontal layered resistivity anomalies.}
\label{GRF_models}
\end{figure}

The GRF models were mapped onto a mesh of 136 $\times$ 50 grids (excluding 10 layers in air with resistivity $10^{10}$ $\Omega$ m for the TE mode). In the central part of the mesh, the cell size is 2 m. The thickness of the first layer is 1 m, and the thickness of the underlying layers increases by a factor of 1.1. Additionally, 10 grids were padded in each horizontal direction with cell size increasing rapidly. The background resistivity value was set as the average of the GRF resistivity values in the core domain. To ensure smooth transitions from the central to the boundary domain, resistivity values were interpolated between these regions. Four randomly chosen models are shown in Fig \ref{GRF_models}. Apparent resistivity and impedance phases for the TE and TM modes were computed for 13 logarithmically spaced frequencies between 1 and 250 kHz at 21 stations covering a 200 m long profile.





\begin{figure}[!h]
\centering
\includegraphics[width=\columnwidth]{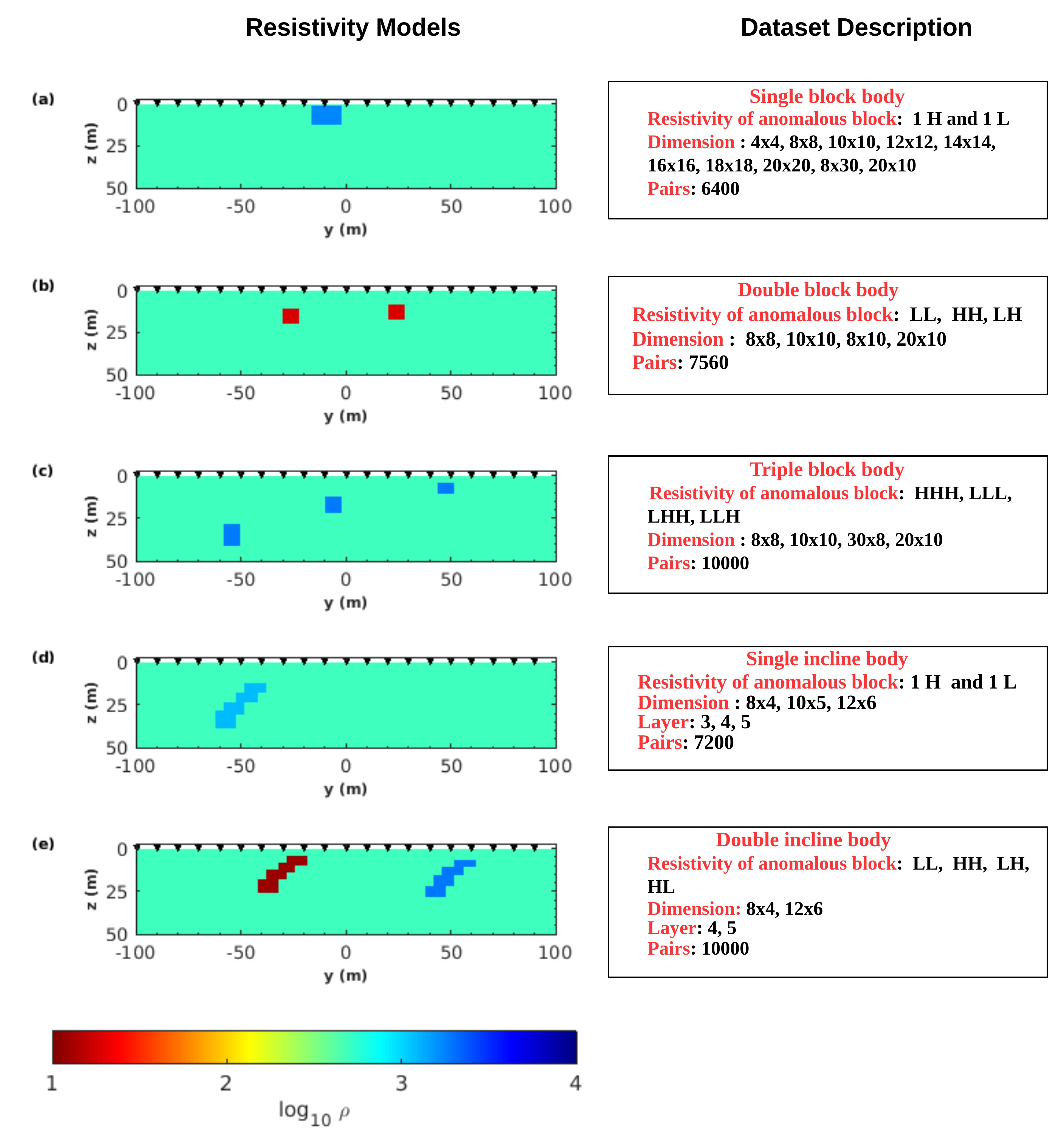}
\caption{Image showing model parameters for generating OOD dataset. The dataset consists of five different types of geometry. (a) Dataset of type-1: single block-shaped anomaly. (b) Dataset of type-2: double block-shaped anomalies. (c) Dataset of type-3: three block-shaped anomalies. (d) Dataset of type-4: single inclined shaped anomaly. (e) Dataset of type-5: double inclined shaped anomaly. The resistivity values for the anomalies are randomly generated in two different classes of resistivity, first \textbf{H} stands for high resistivity between (1000-2000) $\Omega~m$, and second one \textbf{L} stands for low resistivity which ranges between (10-20) $\Omega~m$ and the background is homogeneous with 500 $\Omega~m$. Each geometry is generated in a different amount the highest being the more complex geometry (c) and (e), while the lowest being the simplest among all (a).} 
\label{fig:dataset_representation}
\end{figure}
\subsection{Out-of-Distribution Samples with Distribution shift from the GRF dataset}
\begin{figure}[!h]
    \centering
    \includegraphics[width=\columnwidth]{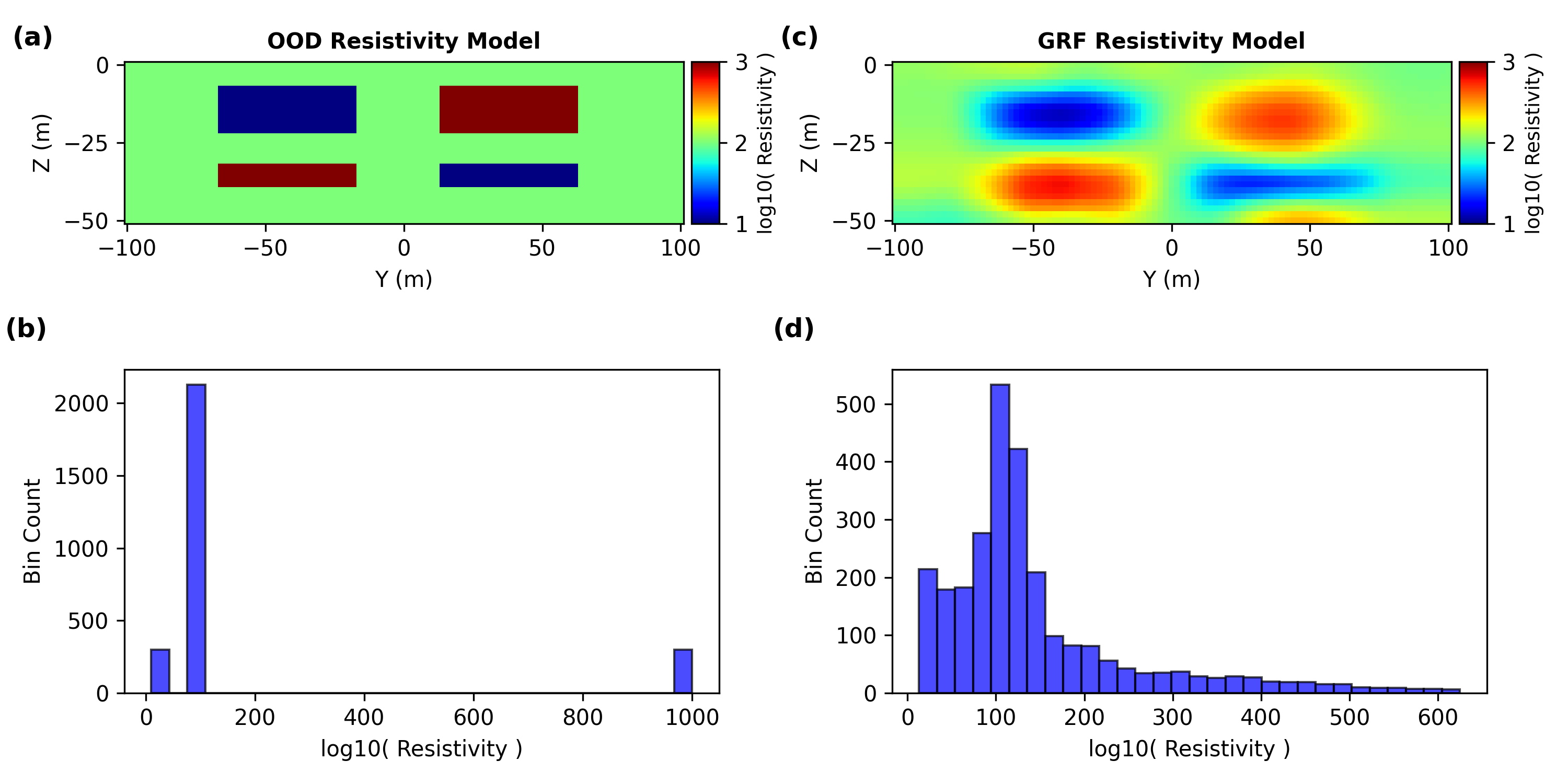}
    \caption{The statistical distribution shift between the checkerboard and GRF resistivity models. (a) represents the checkerboard sample, (b) is the corresponding histogram plot which consists of three peaks denoting background, low and high resistivity, (c) the GRF sample and (d) shows a distribution of GRF sample from high to low resistivity.}
    \label{fig:checkerboard_distribution}
\end{figure}

To evaluate the performance of the proposed U-Net for OOD samples, we have used another dataset \cite{liu2020deep}, henceforth known as the OOD dataset. The dataset consists of rectangular-shaped structures of varying sizes embedded in a homogeneous background. The resistivity for the conductive and resistive structures varies between 10 - 50 $\Omega~m$ and 1000 - 2000 $\Omega~m$ respectively. The synthetic models were subdivided into five different types based on the number, geometry and the associated resistivity values of the anomalous bodies. A schematic representation and description of the various types of models is shown in Fig. \ref{fig:dataset_representation}.  This data is statistically different from the GRF data. The shift in the statistical distribution between OOD samples and GRF samples for a representative sample is shown in Fig. \ref{fig:checkerboard_distribution}. The OOD samples consist of rectangular-shaped resistive and conductive blocks embedded in a homogeneous background. Such kind of model representation results in three peaks in the histogram plot Fig. \ref{fig:checkerboard_distribution}(a) and (b). The plot peaks at three different resistivity values corresponding to the resistivity of conductive and resistive blocks and the resistivity of the background. For the GRF samples in Fig. \ref{fig:checkerboard_distribution}(c), the conductive and resistive bodies follow a Gaussian distribution and are multi-modal in nature where the peak values of the distribution correspond to the resistivity values for the blocks and the background. The Gaussian distribution ensures smooth variations in the resistivity across the model domain, as shown in Fig. \ref{fig:checkerboard_distribution}(d). 

\section{Numerical Results}
\label{result}
\subsection{Model Training} \label{model_training}
\begin{figure}[!h]
\centering
\includegraphics[width=\columnwidth]{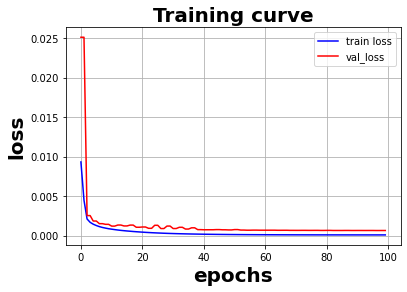}
\caption{Loss curve for U-Net on training and validation set.}
\label{Train_curve}
\end{figure}

\begin{figure*}[!h]
\centering
\includegraphics[width=1.8\columnwidth]{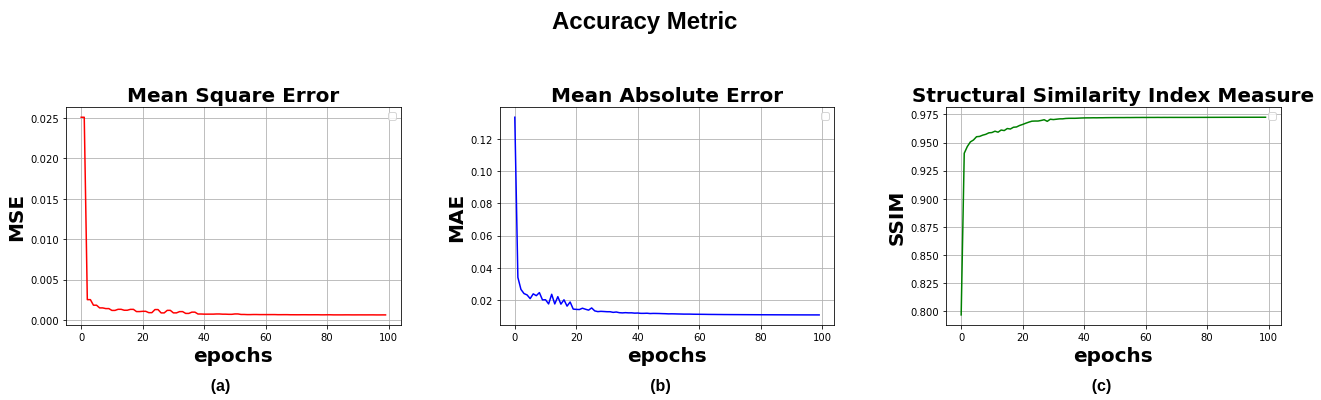}
\caption{Accurac metric on validation samples. (a)Mean Square Error(MSE), (b)Mean Absolute Error(MAE) and (c)Structural Similarity Index Measure(SSIM) scores on the validation dataset.}
\label{Accuracy_curve}
\end{figure*}

\noindent To evaluate the performance of the proposed network, we utilized TorchMetrics, a popular library for metric calculation in PyTorch.  We used MSE and Mean Absolute Error (MAE) to compare the predicted resistivity with the true resistivity models. For structural similarity comparison, we employed the Structural Similarity Index Measure (SSIM). The dataset consists of 41,000 data and model pairs. We allocated 10\% (4,100 samples) of the dataset for testing, with the remaining data split into 76.5\% for training and 13.5\% for validation. The Adam optimizer \cite{kingma2014adam} was used for network training with hyperparameters $\beta_1$ and $\beta_2$, set to 0.9 and 0.999, respectively. The learning rate was set to 0.0001 and managed using the LambdaLR scheduler, which decreases the learning rate by 5\% per epoch. To prevent overfitting, we implemented early stopping with a patience of 20 epochs, guided by the average validation loss and MAE. All computations were performed on an Nvidia DGX cluster. The training process took approximately 68 hours, with convergence achieved after 40 epochs. Figure \ref{Train_curve} shows the training and validation curves, while Figure \ref{Accuracy_curve} illustrates the performance of the U-Net training through accuracy metrics. The SSIM metric approached a value of 0.970, indicating a high similarity between the true and predicted resistivity distributions.
 
\begin{table}[!h]
\centering
\caption{Performance of U-Net with various dataset (both ID and OOD dataset). The $\uparrow$ shows the higher the value better the performance. The $\downarrow$ shows that the lower the values better the performance.} 
\label{tab:metric}
\begin{tabular}{ |c|c|c|c|c|} 
\hline
\multicolumn{5}{|c|}{U-Net} \\
\hline
\multicolumn{2}{|c|}{Dataset} & \multicolumn{3}{c|}{Accuracy metric} \\
\hline
Train Data & Test Data  &  MSE $\downarrow$  &  MAE $\downarrow$  &  SSIM $\uparrow$ \\ 
\hline
GRF & GRF  &  \textbf{0.0006}  &   \textbf{0.010}  &  \textbf{0.970}   \\
\hline
GRF & Blocky &  0.0020  &   0.017  &  0.460   \\
\hline
Blocky & GRF &  0.5050  &   0.277  &  0.133   \\
\hline
\end{tabular} 
\end{table}

\subsection{RMT data inversion for IDD samples}
\begin{figure*}[!h]
\centering
\includegraphics[width=2.0\columnwidth]{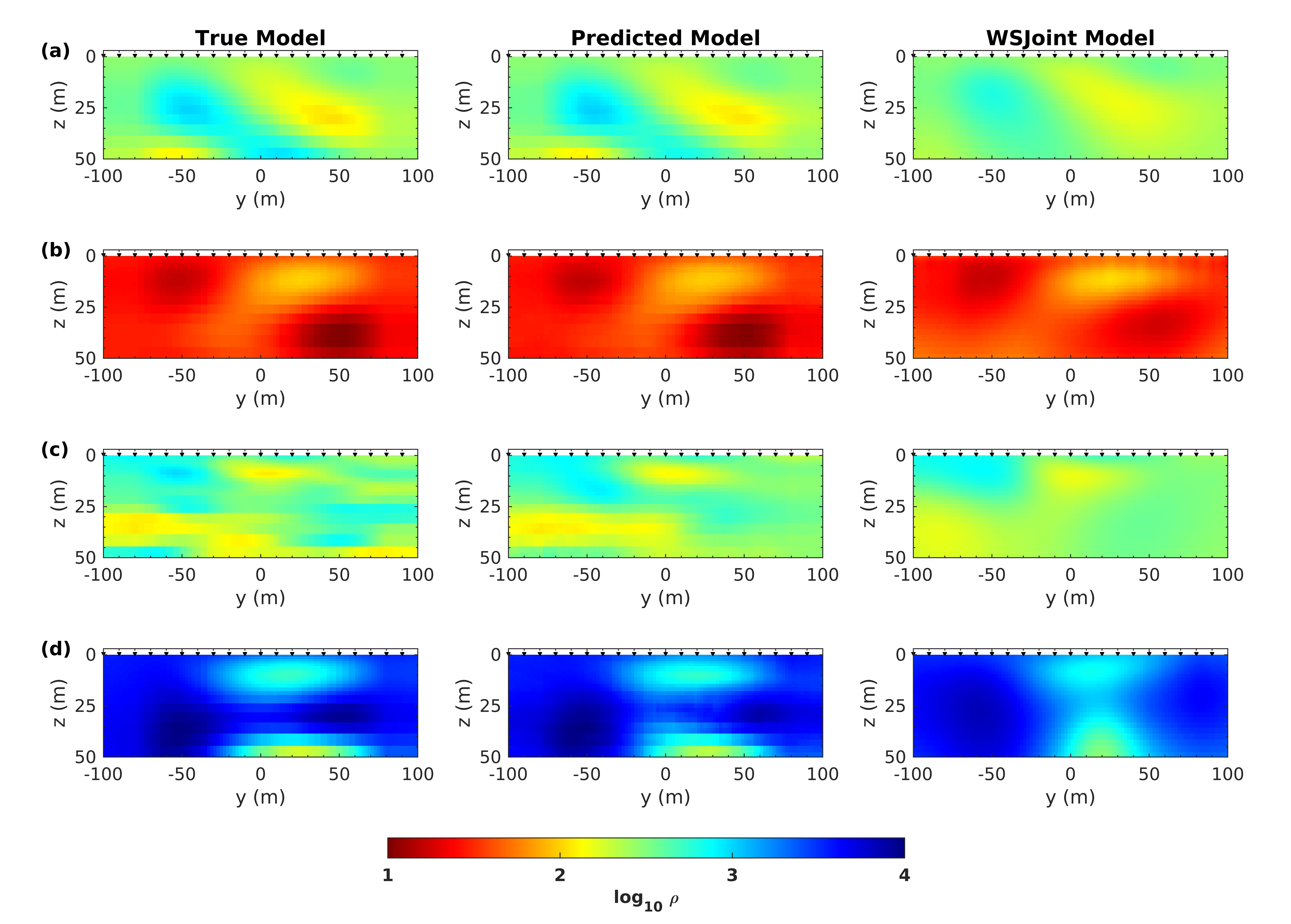}
\caption{Image showing inversion results using U-Net for 4 randomly chosen samples from the IDD test dataset. The three columns starting from the left show the true resistivity model, the inverse model using U-Net, and the inverse model obtained using WSC. The RMT sites are shown as black triangles at the top of each plot.}
\label{exp1}
\end{figure*}

To evaluate the performance of the GRF-trained network, we randomly selected four samples from the IDD test dataset and performed inversions using the U-Net. Additionally, we conducted traditional inversions for each dataset using WSJointInv2D-MT-DCR code \cite{amatyakul2017wsjointinv2d}, henceforth known as WSC. For the inversion using WSC, the grid is identical to one used to generate the synthetic data, and the starting model was set to the mean of TE and TM mode apparent resistivity. The inversion results, alongside the true models, are presented in Figure \ref{exp1}. Across all four samples, the U-Net successfully recovered the resistivity models, with the predicted models closely resembling the true models. For instance, in the U-Net inversion results shown in Figure \ref{exp1}(a), the two resistive structures with resistivities of $10^3$ and $10^2$ $\Omega m$ are accurately delineated in terms of shape, size, and associated resistivity value. In the half-layer of resistivity 90 $\Omega m$ embedded in the resistive background, the structure is imaged, although with less complexity as compared to the true model (see Figure \ref{exp1}(c)). In Figure \ref{exp1}(d), the two conductive anomalies within a resistive background are correctly imaged, and the separation between the two structures is clearly visible. In contrast, the inversion results using the WSC code show some dissimilarities. As seen in Figure \ref{exp1}(a), the shape of the resistive body is not properly delineated, with structures becoming smeared beyond a depth of 30 m. The smearing is due to the model regularisation implemented in the  WSC. In Figure \ref{exp1}(c), the small conductive body beneath the top resistive body (as compared to the background) is not imaged, although the top resistive body is well captured. In Figure \ref{exp1}(d), the resistivity layer sandwiched between two conductive layers is missing, and the top and bottom conductive anomalies appear to be connected. Overall, the U-Net demonstrates superior performance in recovering the resistivity models compared to the traditional deterministic inversion algorithm. The quantitative analysis for the complete test dataset is summarized in the first row of Table \ref{tab:metric}. The low MSE indicates a minimal discrepancy between the true and predicted resistivity models. Additionally, high correlation metrics and SSIM values confirm the structural coherence between the predicted and true models The average misfit is about 3$\times 10^{-4}$ $\Omega$m and the maximum misfit is less than 6$\times 10^{-4}$, which can be considered a very small misfit in the context of this study. 

\subsection{RMT data inversion for ODD samples}
\begin{figure*}[!h]
\centering
\includegraphics[width=1.8\columnwidth]{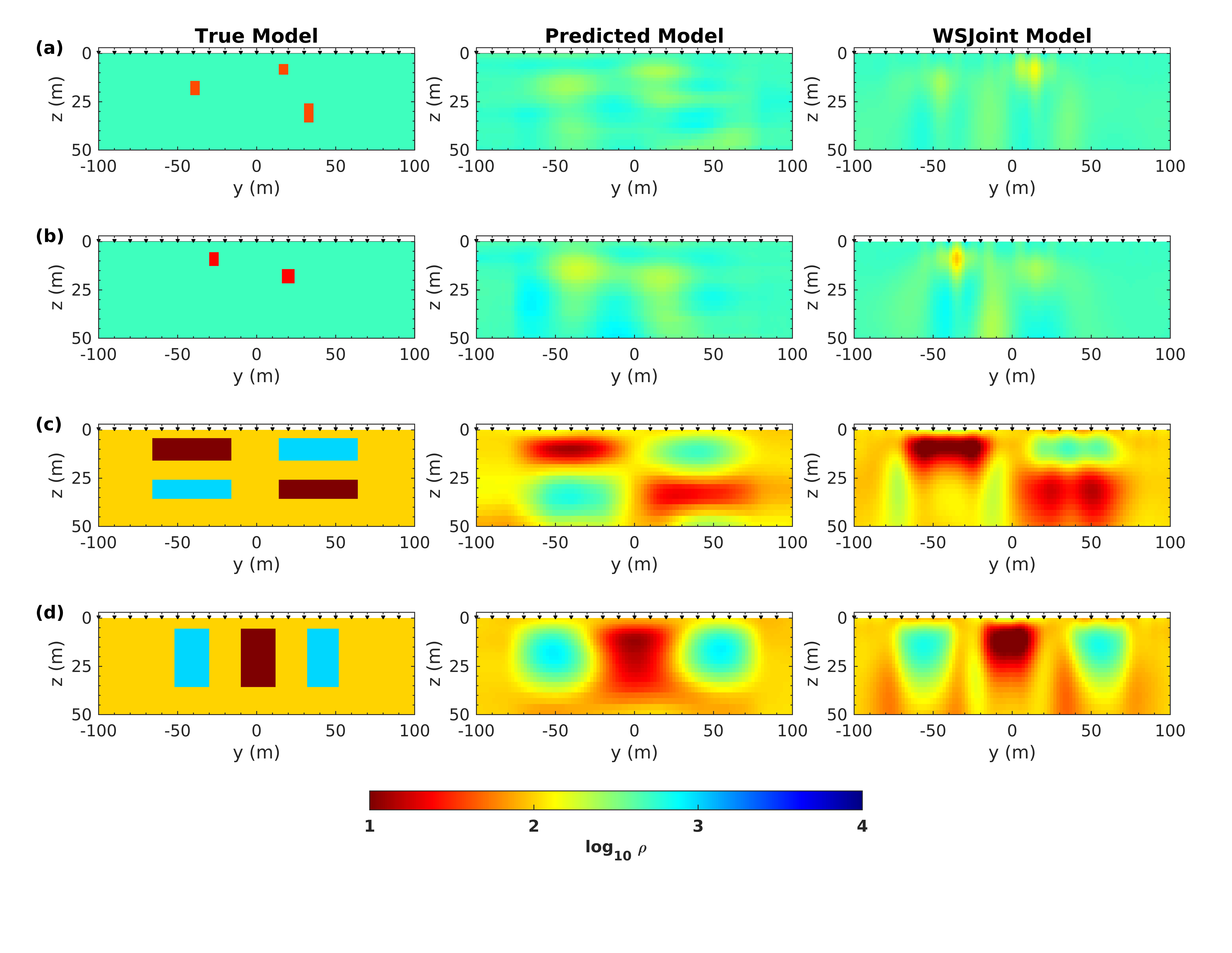}
\caption{Inversion results using U-Net for 2 randomly chosen samples from the ODD and 2 checkerboard models. The firts, second, and third columns show the true model, the inverse model using U-Net, and the inverse model using WSC respectively. The RMT sites are shown as black triangles.}
\label{ood1}
\end{figure*}

To test the performance of the U-Net on the ODD dataset, we selected 2 random models from the ODD test data and created 2 checkerboard models. The first ODD sample contains two resistive and one conductive anomalies placed in a 500 $\Omega \text{m}$ half-space, while the second ODD sample contains two conductive and two resistive anomalies in a similar half-space. The first checkerboard model features three blocks of dimension 30 $\times$ 25 embedded in a 100 $\Omega \text{m}$ background, with the top of the blocks at z = 5 m. The central block is conductive (100 $\Omega \text{m}$), flanked by two resistive blocks with a resistivity of 1000 $\Omega \text{m}$. In the second checkerboard model, two conductive blocks (10 $\Omega \text{m}$) and two resistive blocks (1000 $\Omega \text{m}$) are embedded in a 100 $\Omega \text{m}$ half-space. The top two blocks are located between (-60, -20) m and (20, 60) m in the horizontal dimension, whereas (5, 15) m in the vertical dimension. The bottom two blocks are located between (25, 35) m in vertical dimension and their horizontal dimensions remain the same as the top two blocks. Figure \ref{ood1} presents the true model (first column), and inverted models obtained using the U-Net (second column) and WSC (third column). 

For the 2 ODD models, the U-Net successfully identified the conductive bodies, with their positions closely matching those in the true resistivity model. However, the network struggled to accurately delineate the shape of these bodies. This challenge arises because, in the synthetic models generated using the Gaussian Random Field GRF, the anomalies are generally large in size and the resistivity values have a smooth transition between anomaly and the background. Models with small anomalies in terms of size are rare, and when present, they typically form part of a more complex resistivity distribution. For the same ODD dataset, the WSC recovers only the shallow blocks, while deeper features are poorly resolved. For instance, in Figure \ref{ood1} (b), the deeper conductive block is not accurately mapped due to reduced sensitivity and the smoothing constraints imposed through model regularization in the WSC.

For the first checkerboard model, the shape and associated resistivity values are very well imaged by the U-Net, although the size of the block appears slightly larger than the true size. For the second checkerboard model, all the four blocks are delineated. However, minor deviation is observed in the size of the blocks. This is expected due to the smooth transition between anomalous zones and the background. The WSC recovers all three blocks in the first checkerboard model but struggles with the second checkerboard model. While it images the two conductive blocks, the base of the bottom conductive blocks is smeared. Additionally, the top resistive block's shape is not accurately delineated, and the bottom resistive block is not imaged at all. In other test examples (not shown here), we observed that the U-Net outperforms the WSC algorithm in capturing complex and deeper resistivity distributions.

\subsection{Experiment with noisy data}
\begin{figure*}[!h]
\centering
\includegraphics[width=1.8\columnwidth]{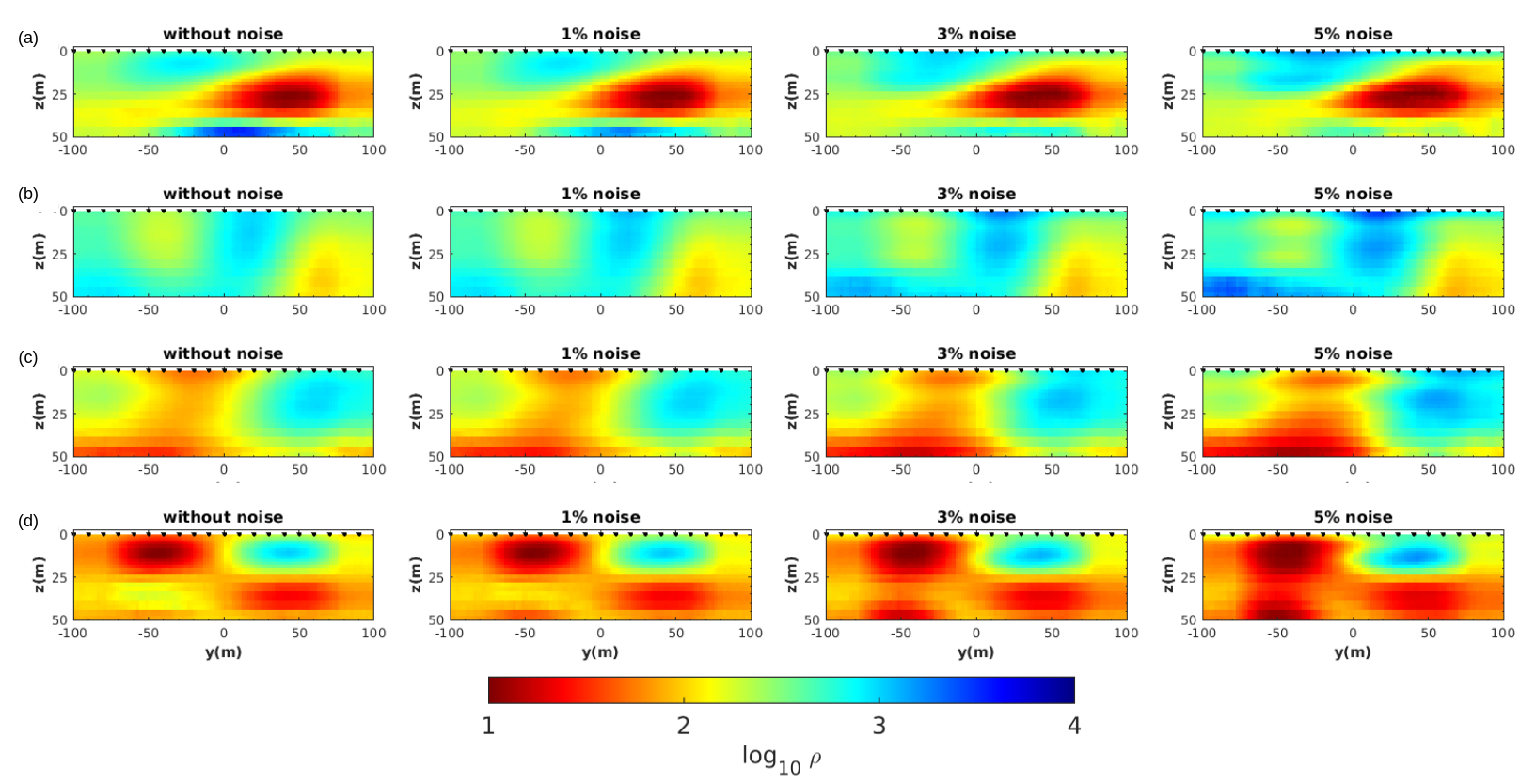}
\caption{Image showing the impact of noise on U-Net inversion for (a-d), using randomly selected models from the test data. The first column displays the inversion results for noise-free data. The second, third, and fourth columns show the inversion results for datasets with 1\%, 3\%, and 5\% noise, respectively. RMT sites are represented as black circles at the top of the plots.}
\label{noise_fig}
\end{figure*}

Noise is an inevitable part of RMT field data. To evaluate the robustness of U-Net-based inversion, we selected datasets for four different models and created new datasets by adding 1\%, 3\%, and 5\% Gaussian noise to the original test data. The inversion results for these noisy datasets are shown in Figure \ref{noise_fig}. From the figure, we observe that for the 1\% noisy data, the inversion results are almost identical to those obtained with noise-free data. However, as the noise level increases, the quality of the inversion deteriorates, particularly at greater depths. For instance, in Figure \ref{noise_fig}(a), with increasing noise, the smaller high resistivity anomalies at the shallow and deeper parts of the model become more deformed compared to the larger conductive anomaly. Similar behavior is observed in Figures \ref{noise_fig}(b-c), where the resistive anomalies become harder to delineate as noise in the dataset increases. In Figure \ref{noise_fig}(d), the bottom resistive part merges with the top conductive block and is completely missed as the noise level increases. This can be attributed to the fact that the response of deeper features at the RMT sites is weak, and it is even weaker for resistive structures. Hence, deeper resistive structures become rapidly distorted with increasing noise. A similar effect is seen for small features at shallow depths. Features with a large response are affected last. For example, the shape of the large conductive body remains consistent even as the noise increases from 1\% to 5\%.

\subsection{Expriment with compressed sensing}

\begin{figure*}[!h]
\centering
\includegraphics[width=1.8\columnwidth]{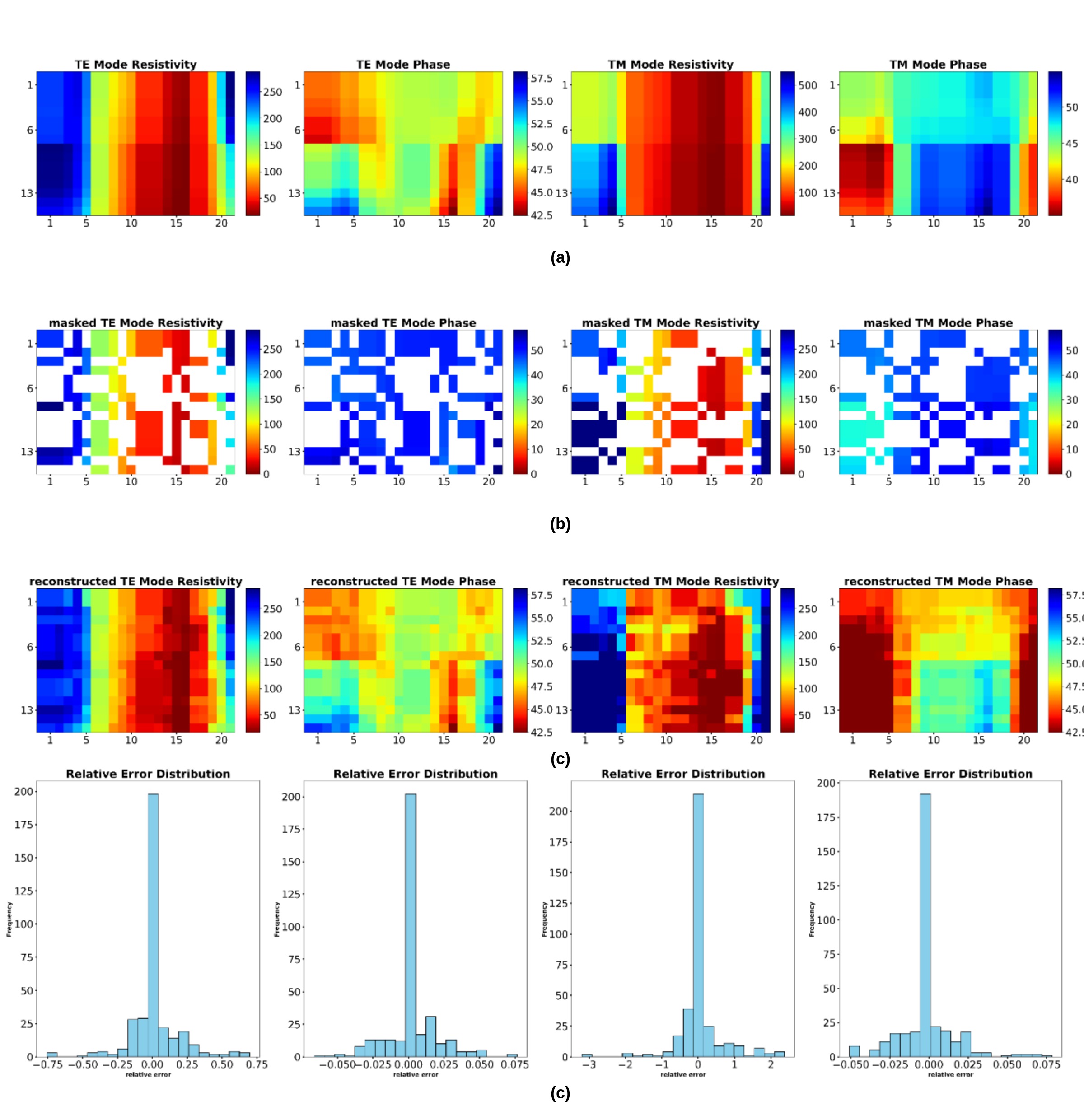}
\caption{Comparison of reconstructed data using compressed sensing and the original dataset. (a) original data, (b) masked data where data points are removed randomly, (c) reconstructed data using the CS approach, and (d)  histogram showing relative error in reconstructed data with respect to the original data.}
\label{exp6}
\end{figure*}

\begin{figure}[!h]
\centering
\includegraphics[width=\columnwidth]{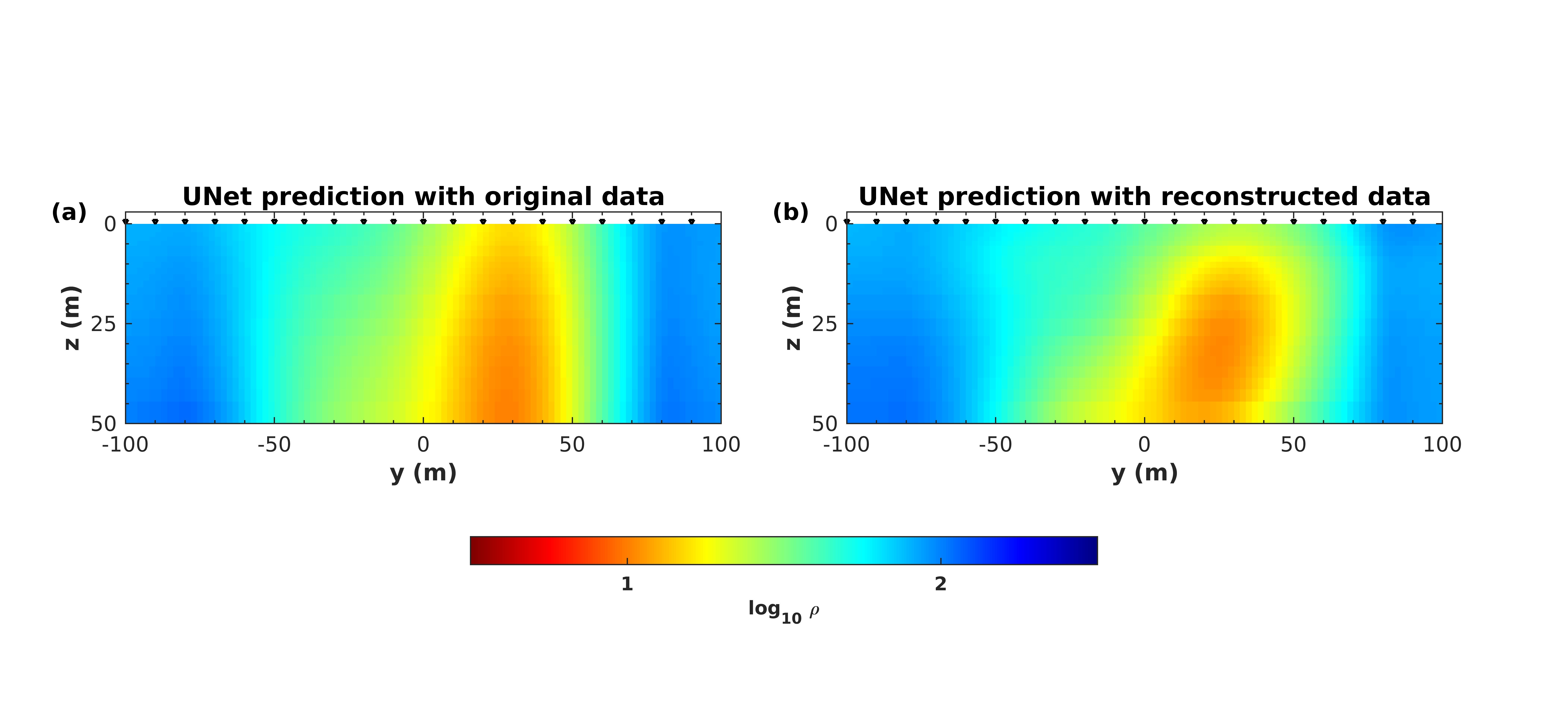}
\caption{Comparison of the inverted model from the reconstructed image and the original image. (a) U-Net predicted model from the original RMT data without any missing values and (b) U-Net prediction for reconstructed data created by masking out values from the RMT data}
\label{exp7}
\end{figure}

For estimating the missed data, we have used the L1 regularized least squares condition with a convex optimization strategy \cite{candes2008introduction}. To validate the CS approach for the RMT data, two experiments were performed. In the first experiment, we examine the quality of the reconstructed RMT data. To achieve this, first, we randomly mask the portion (31.25\%) of RMT data and then apply the CS scheme to reconstruct the original data. The original, masked and the reconstructed data is shown in Figure \ref{exp6}(a-c). The four columns in the image represent each channel of the RMT data. To assess the quality of the reconstructed we computed the relative error between the original and reconstructed data as
\begin{equation}
    \text{relative error} = \frac{\text{reconstructed data} - \text{original data}}{\text{original data}}
    \label{rel_err}
\end{equation}
Figure \ref{exp6}(d) shows the relative error in terms of the histogram for each channel. The relative error is small and the reconstructed image is similar to the original image.  However, a relatively high error is also observed. The high relative error can be attributed to a large masked contiguous portion in a data channel. The quality of the reconstruction degrades as the degree of masking increases in the dataset. In the second experiment, we use the reconstructed data as the input for the U-Net inversion and compare the inversion results against those obtained using the original data. Fig\ref{exp7} shows U-Net inversion with original data and with the reconstructed data. The two inverted models are nearly the same, representing a low-conductive body embedded in a resistive host. Both experiments validate the utilization of CS for RMT data. However, as the masking percentage increases, a deterioration in the quality of inversion results is expected.

\section{Field example}
\label{field_section}
\begin{figure*}[bt]
\centering
\includegraphics[width=1.8\columnwidth]{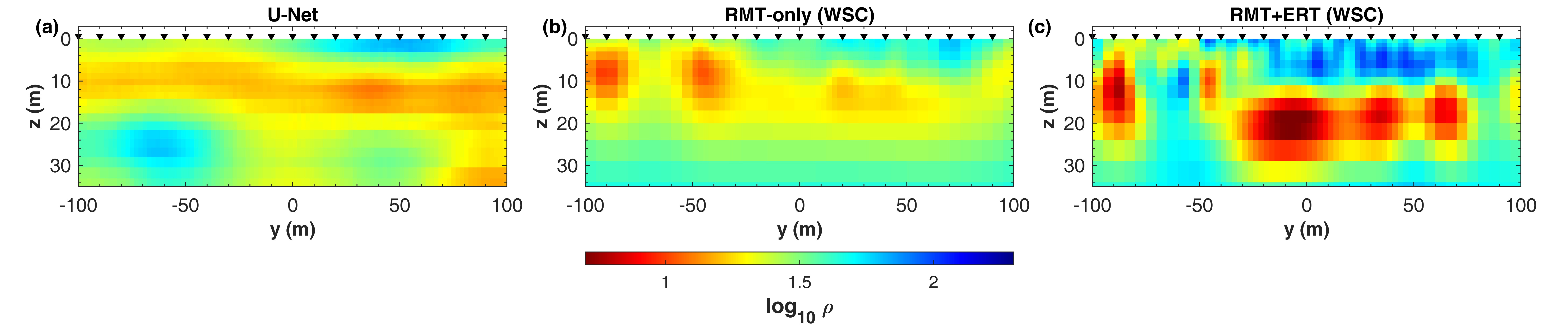}
\caption{Image showing inverse model obtained using (a) U-Net, (b) RMT-only inversion using WSC (c) joint inversion of RMT and ERT data using WSC.}
\label{field}
\end{figure*}

To validate the developed scheme, we inverted an exemplary dataset from Saliyar village near Roorkee, India. This data was originally acquired by \cite{yogeshwar2012groundwater} to investigate groundwater contamination due to seepage from a waste disposal site. A total of 13 RMT profiles, including a reference profile, were acquired using an RMT-F instrument from the University of Cologne \cite{kirsch2006groundwater}. On average the profile length is about 200m long with 10 m of station spacing. For TE and TM modes, four to six frequencies were available in the range of 10kHz to 1MHz. The authors also acquired electrical resistivity (ERT) data using wenner-schlumberger configuration. For the ERT data, 41 electrodes were placed along the profile with inter-electrode spacing of 2.5 m in the central part of profile and 5 m on either side. The reference profile is far from the contamination area and lies in the Solani river bed. In this part of the survey area, the uppermost layer comprises sandy loam, characterized by a thickness spanning 3 to 6 meters. This layer showcases a fairly uniform distribution of soil, sand, silt, and clay \cite{sin_roy_2003}. Beneath this lies a shallow unconfined aquifer, extending from 3 to 27 meters in thickness, succeeded by a second aquifer with an approximate thickness of 14 meters. \cite{yogeshwar2012groundwater} had previously performed 2D inversion of the recorded RMT data. Along the reference profile, a few data points were noisy and were discarded during data processing. To reconstruct the missing data, we employed the CS approach. Once the dataset was reconstructed, we performed inversions using the U-Net and WSC. For the inversion using WSC, a 50 $\Omega$-m half-space was used as the starting model. The grids used for model discretization were similar to those used for generating the synthetic data. Following Devi et al. (2020), the error floor was set to 5\% for apparent resistivity and 2.5\% for the phase. The inversion results are shown in Figure \ref{field}.


Two main features of the reference profile are the top unsaturated resistive
layer (60-150 $\Omega$-m) overlying the water-saturated conductive layer (30 $\Omega$-m). In the inverted model obtained using the U-Net, the resistive feature is visible and limited to the south of the profile in the depth range of 0-6 m. The conductive water-saturated layer is visible from the surface to a depth of 25 m. Additionally, a high resistive zone is observed between depths of 20-30 m. In the inverted model using WSC, the top unsaturated layer is present in the southern part of the profile, and the water-saturated conductive layer extends to a depth of 25 meters, with its base smearing into the background. However, the resistive zone at the base of the conductive layer is not present in this model. Furthermore, using WSC, we performed the joint inversion of the RMT and ERT data. In the joint inversion model, the top resistive layer extends to a depth of 10 meters, and a resistive zone is observed at depths of 20-30 meters between -70 to -50 meters. This resistive zone is similar to the one observed in the inverted model obtained using the U-Net. The ability of the U-Net to image the deeper resistive zone, imaged by the deterministic inversion only in the joint inversion framework, demonstrates the robustness of the proposed network.
To further validate the inverse model, we show the data misfit for four representative sites. The forward solutions for the inverted model obtained using the U-Net were computed using the FD based solver. The data are well fitted although at few places we observe a slight mismatch between the observed and the predicted data (for example, in the TE mode apparent resistivity for site number 09). The fit is good for the higher periods as compared to the lower periods.

\begin{figure*}[bt]
\centering
\includegraphics[width=1.5\columnwidth]{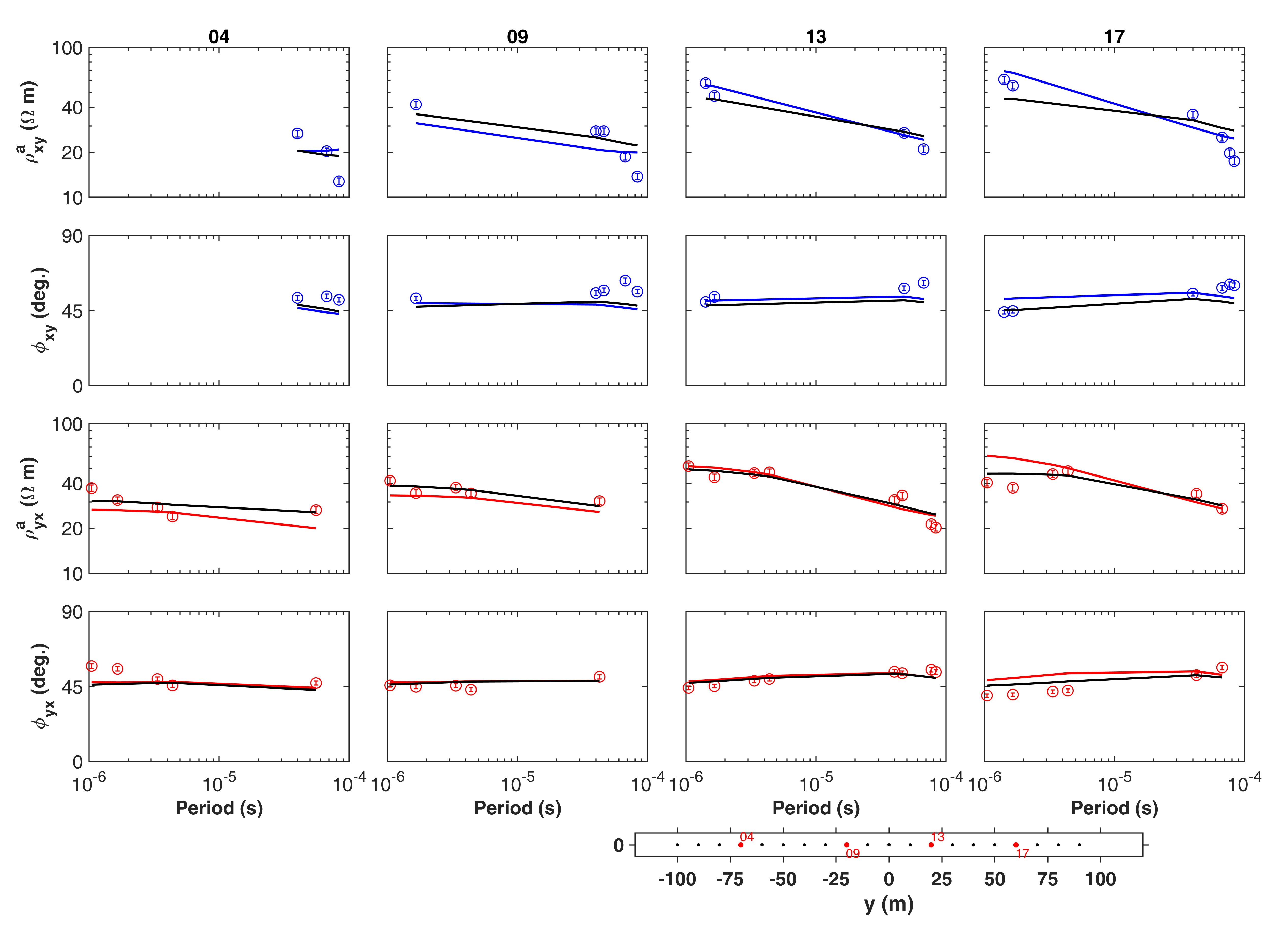}
\caption{The misfit plot of RMT data in terms of apparent resistivity and phase versus periods. Red/blue circles represent the observed data, red/blue lines represent predicted data for the model obtained using the U-Net. The predicted data corresponding to the inverted model obtained using WSC is illustrated by the black line. The location of the RMT sites is shown at the bottom of the image.}
\label{field_data}
\end{figure*}
\section{Conclusion}
\label{conclu}

In this study, we developed a scheme for 2D RMT data inversion using deep learning. To increase the network's generalization we created the test data using GRFs and modified the existing U-net. 
The developed scheme was tested over different models randomly taken from the test data. The performance of the proposed network exceeds that of a deterministic inversion scheme even for the ODD samples. The proposed scheme helps to overcome the problem of retraining or fine-tuning the network over additional data when applied to the field. High accuracy on the OOD and as well as checkerboard models shows the efficiency of the network on unseen data with statistical distribution shift. The robustness of the network is checked by evaluating the results in the presence of different amounts of noise levels. In each of these experiments, the proposed scheme surpasses the WSC in terms of recovery of resistivity values and resistivity structures.
To deal with the missing data point CS scheme was deployed. The quality of the reconstructed data using the CS scheme is comparable to that of the original data. The scheme was tested over field data from ground water contamination site from Solani, Roorkee, India. The inversion of the field data reveals features that were not mapped using deterministic inversion. The generalization error can be further reduced by increasing the amount of data and using a more advanced network or using a physics-based simulator as a loss function to guide the training process to accurate results.

\newpage
\section*{Acknowledgments}
\noindent The authors are grateful to the High Computer Center, IIT (ISM) Dhanbad for providing the computational facility. The authors would like to thank Mohammad Israil for providing the field data.

\section{References Section}
\bibliographystyle{IEEEtran}
\bibliography{sample}
\newpage
\onecolumn

\vfill

\end{document}